\documentclass[10pt,twocolumn,letterpaper]{article}

\usepackage{times}
\usepackage{epsfig}
\usepackage{graphicx}
\usepackage{amsmath}
\usepackage{amssymb}
\usepackage{booktabs}
\usepackage{url}
\usepackage{hyperref}
\usepackage{xcolor}
\title{Revisiting KRISP: A Lightweight Reproduction and Analysis of Knowledge-Enhanced Vision-Language Models}
\author{ 
\begin{tabular}{cc} 
\begin{tabular}{c} 
\textbf{Souradeep Dutta} \\ \textit{Koita Centre For Digital Health} \\ Indian Institute of Technology Bombay \\ Mumbai, India \\ souradeep.dutta@iitb.ac.in 
\end{tabular} 
& 
\begin{tabular}{c} 
\textbf{Keshav Bulia} \\ \textit{Department of Metallurgical engineering} \\ \textit{\& Material science} \\ Indian Institute of Technology Bombay \\ Mumbai, India \\ 21D110008@iitb.ac.in \end{tabular} \\[1.2em] \multicolumn{2}{c}{ \begin{tabular}{c} 
\textbf{Neena S Nair} \\ \textit{Department of Bioscience \& Bioengineering} \\ Indian Institute of Technology Bombay \\ Mumbai, India \\ 24m0153@iitb.ac.in 
\end{tabular} } 
\end{tabular} }

\begin{document}
\date{}
\maketitle
\begin{abstract}
Facebook AI Research introduced KRISP \cite{marino2020krisp}, which integrates structured external knowledge into pipelines for vision-language reasoning.  Despite its effectiveness, the original model has been developed for industrial-scale training, is computationally demanding, and is tightly connected to a large backbone.  In this work, we reexamine KRISP from a different angle and offer a lightweight reproduction with significantly fewer parameters.  Even though our replicated model performs about 75 \% of the original, the replication process uncovers a number of design flaws, real-world pitfalls, and implicit problems that were not fully covered in the original paper.  We offer insights into the scalability and efficacy of knowledge-enhanced VQA architectures under resource constraints through systematic ablation studies, which include a proof-of-concept on synthetic VQA data and evaluation on the DAQUAR dataset.

\textbf{Our model, configured with a low parameter setup and constrained by the external Knowledge graph domain, prevents AI hallucinations and generates outputs solely within that domain. Minimal parameters allow us to function on edge devices like smartphones and AR-VR, further improving offline visual reasoning.}

\end{abstract}

\section{Introduction}
Knowledge-enhanced vision language models have become an important tool for enhancing reasoning beyond information based solely on images. Facebook AI Research proposed KRISP, \cite{marino2020krisp} which combines neural embeddings and deterministic knowledge graphs to allow explicit grounding of entities during visual question answering (VQA)\cite{wang2017fvqa}. Despite its advantages, KRISP is still computationally costly and difficult to replicate without substantial resources.

This work is a course-project-level re-examination of KRISP, focusing on:\vspace{-4pt}
\begin{enumerate}
    \item Understanding its architectural dependencies,
    \item Diagnosing practical shortcomings, and
    \item Exploring whether a significantly smaller model can replicate its reasoning ability.
\end{enumerate}

We reimplement a minimally parameterized version of KRISP with the same conceptual objective of integrating structured knowledge with visual features. In order to test "how much" of KRISP's foundational concept can endure under severe parameter constraints, our models are intentionally small.

\section{Flagged Issues in the Original KRISP Model}
\begin{figure}
    \centering
    \includegraphics[width=1\linewidth]{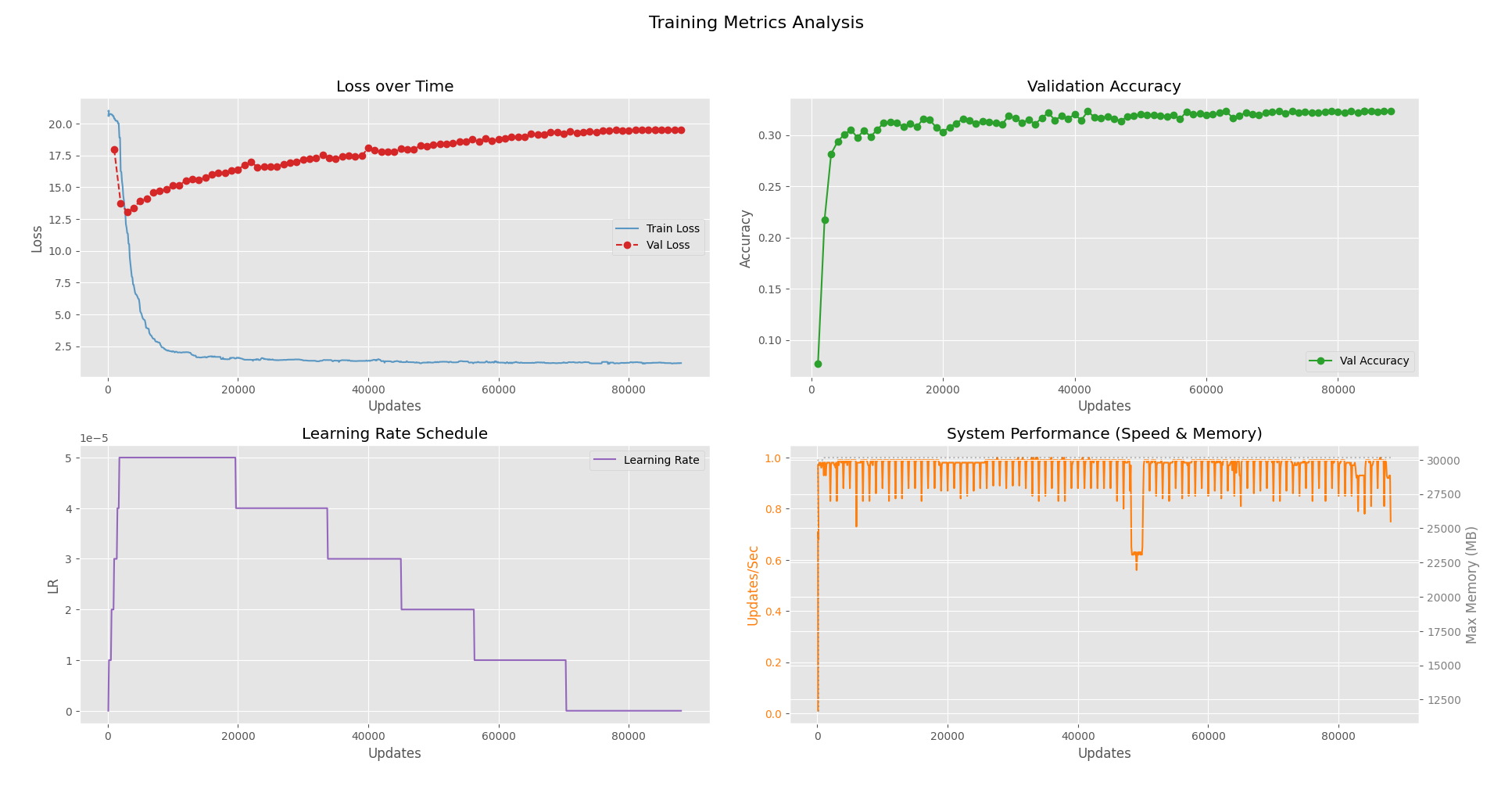}
    \caption{The above image shows the detailed evaluation of the reproduced KRISP architecture on RTX A6000.}
    \label{krisp_eval}
\end{figure}

Despite the KRISP paper's robust conceptual framework, our reproduction efforts uncovered a number of practical problems that the original work does not specifically address. Graphical details further provided to support our claim~\ref{krisp_eval}.To help future lightweight replications, we enumerate these observations in brief below.

\begin{enumerate}

    \item \textbf{Severe overfitting tendencies.} The model is vulnerable to overfitting, particularly when replicated without extensive pretraining, due to the close coupling of visual, linguistic, and knowledge embeddings in the KRISP architecture.
    \item \textbf{High parameter sensitivity.} Performance is severely hampered by slight variations in component dimensionality (such as knowledge projector sizes).
    \item \textbf{Large computational footprint.} Reproducing the original model in academic settings is challenging due to its high GPU memory requirements and lengthy training cycles.
    \item \textbf{Implicit engineering assumptions.} Many implementation details (preprocessing, entity alignment, graph pruning) are not stated clearly and require manual reverse engineering.
\end{enumerate}

\section{Lightweight Reproduction Method}
Our objective was to maintain the fundamental concept of KRISP—combining visual representations with external structured knowledge—rather than to precisely replicate it. As a result, we apply various KRISP-like module variations with aggressive parameter reduction.

\subsection{Core Architecture Components}
For both visual and textual inputs, the corresponding CLIP \cite{radford2021clip} encoder is used as a frozen feature extractor in all models. This is followed by trainable projection layers and attention\cite{vaswani2017attention}-based fusion modules. Among the essential architectural elements are:

\begin{enumerate}
    \item \textbf{Visual and Question Encoding:} Linear layers project CLIP-extracted features (512D)
    \item \textbf{Knowledge Retrieval:} Local ConceptNet\cite{speer2017conceptnet} querying using question keywords and identified image concepts
    \item \textbf{Cross-Attention Fusion:}Multi-head focus between pairs of images and questions
    \item \textbf{Knowledge Integration:}  Knowledge graphs are used to add domain-bound knowledge from outside sources to the model.
    \item \textbf{Answer Prediction:} The final response uses BERT\cite{devlin2019bert} as a language model. 
\end{enumerate}

\subsection{Model Variants and Ablation Studies}

\begin{figure*}[h]
    \centering
     \includegraphics[width=1\linewidth]{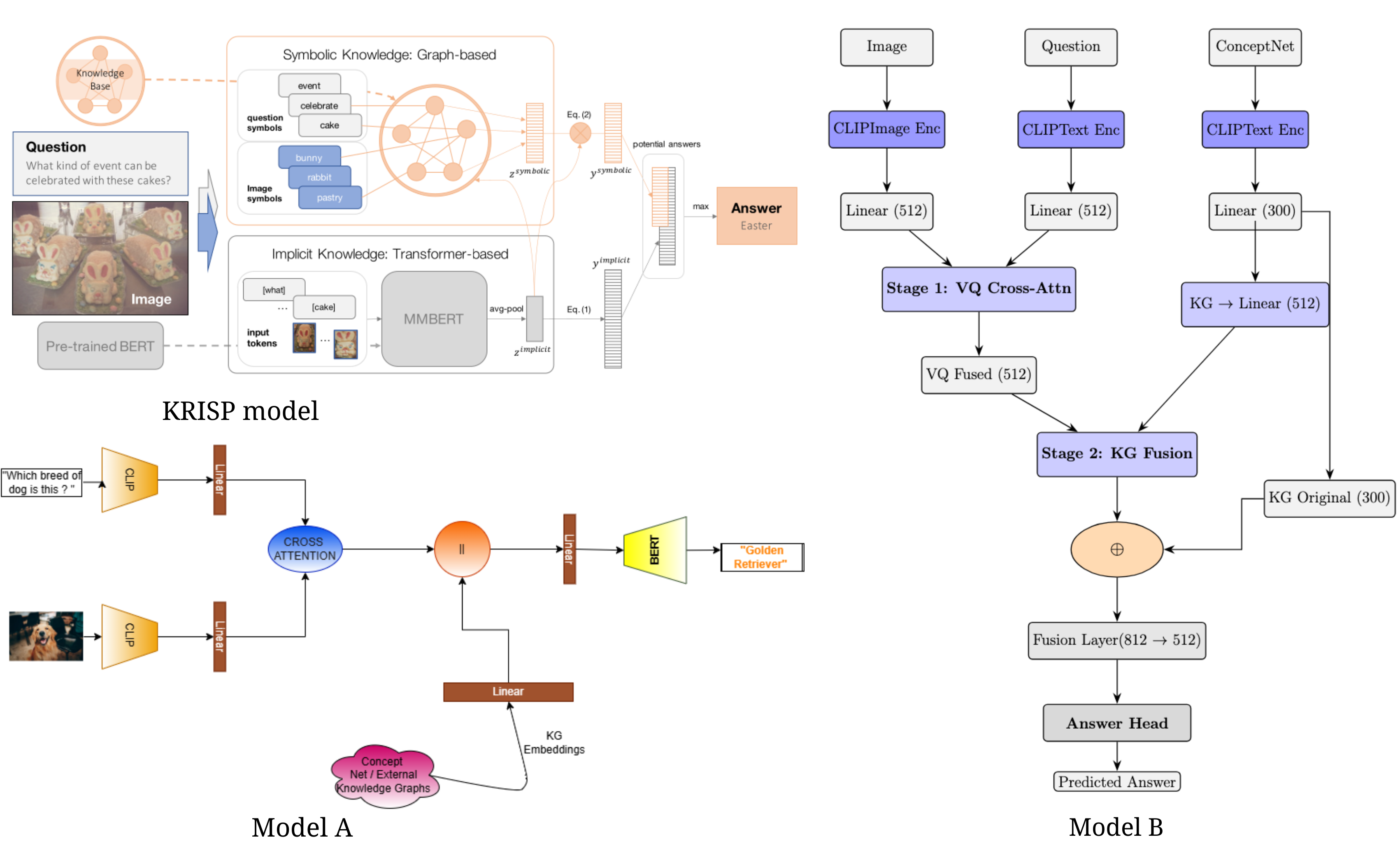} 
    \caption{ Krisp Model Architecture \cite{marino2020krisp} and Proposed Lightweight Architecture Models A and B}

    \label{Combined VQA Architectures}
\end{figure*}

We analyze three progressive variants through ablation studies:

\textbf{Model A (Best Performing):} ModelA tested on VQAV2, achieved approximately 75\% performance of original. A lightweight architecture ~\ref{Combined VQA Architectures} maintaining the core logic of the KRISP structure with reduced dimensions. Uses approximately 22\% trainable parameters of the actual KRISP model , with CLIP embeddings (512D), reduced knowledge embeddings (300D), and 8-head attention.Although it suffers from overfitting too, the validation accuracy it has reached, in contrast to the total parameter count, suggests that the model has potential to be scaled.
\begin{figure}
    \centering
    \includegraphics[width=1\linewidth]{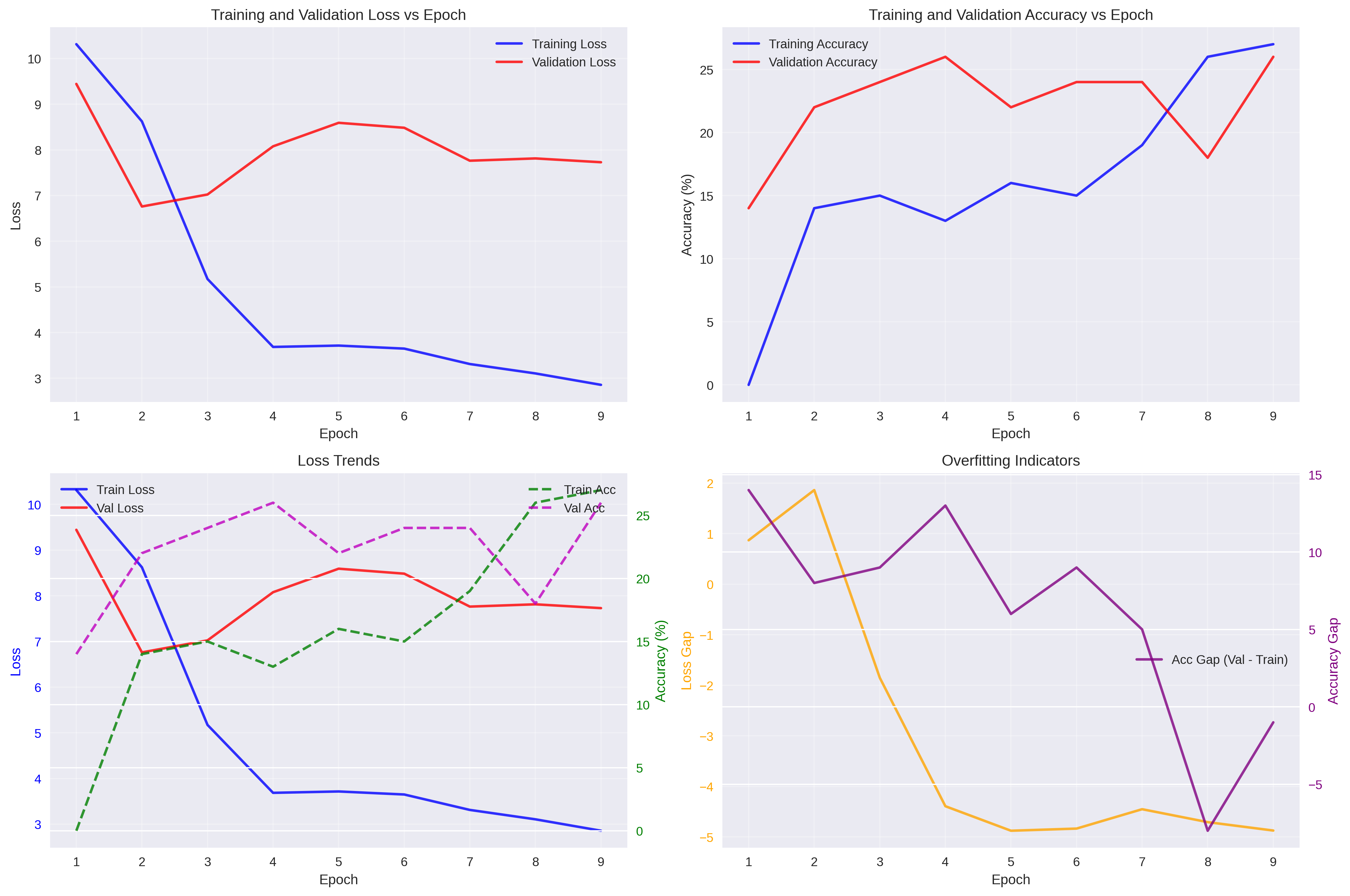}
    \caption{Detailed Model A performance.}
\end{figure}

\textbf{Model B (Modified Architecture):}Building on Model A, we introduced architectural modifications~\ref{Combined VQA Architectures} to address observed limitations. The key difference lies in the knowledge integration strategy: rather than simple concatenation, we employ a two-stage attention mechanism where image-question fusion occurs first, followed by knowledge-grounded refinement. This change was motivated by the observation that direct knowledge concatenation can dilute visual grounding. Parameters: ~3.63M trainable.

\textbf{Proof of Concept for Model B and VQA (DAQUAR DATA):}Final assessment using a real-world indoor scene VQA benchmark, the DAQUAR \cite{malinowski2014daquar} dataset.  DAQUAR includes human-annotated questions about indoor scenes with intricate spatial relationships and object interactions, in contrast to synthetic data.  We improved knowledge retrieval so that it was grounded in images:  ConceptNet retrieves knowledge triples related to \textit{both} detected concepts and question keywords after CLIP's zero-shot classification \cite{radford2021clip} first identifies visual concepts in the image (such as "desk", "monitor", and "keyboard").  There are 6,149 samples in the training set, 5,076 samples in the validation set, and 582 distinct answers in the answer vocabulary.

\section{Detailed Architecture Analysis}

\subsection{Image-Grounded Knowledge Retrieval}
The image-grounded knowledge retrieval mechanism is a key innovation in our methodology. Instead of using question keywords alone to retrieve knowledge (which may introduce insignificant information), we:

\begin{enumerate}
    \item  Utilize CLIP's zero-shot capabilities to identify the top five concepts in the final image embedding.
    \item Identify important entities in the question text.
    \item  Query ConceptNet for triples that contain identified image concepts first (priority).
    \item If necessary, add information relevant to the question. To reduce noise, stick to the top five most pertinent triples.
\end{enumerate}

This grounding lowers the possibility of hallucinations or irrelevant knowledge injection by ensuring that retrieved knowledge is anchored to actual image content.

\subsection{Two-Stage Knowledge Injection Mechanism}
Unlike simple feature concatenation, our architecture employs cascaded attention:

\textbf{Stage 1-Visual-Question Fusion:} A question-aware visual encoding is produced through multi-head cross-attention, in which image features pay attention to question representations.

\textbf{Stage 2: Integration of Knowledge:} The model can then selectively incorporate pertinent external information by fusing the fused visual-question representation to knowledge embeddings.

A common mode of failure in naive fusion strategies is knowledge overwhelming visual grounding, which is prevented by this staged approach.




\section{Results}

\subsection{Quantitative Results}
Table~\ref{tab:results} summarizes performance across model variants and datasets.

\begin{figure}
    \centering
    \includegraphics[width=1\linewidth]{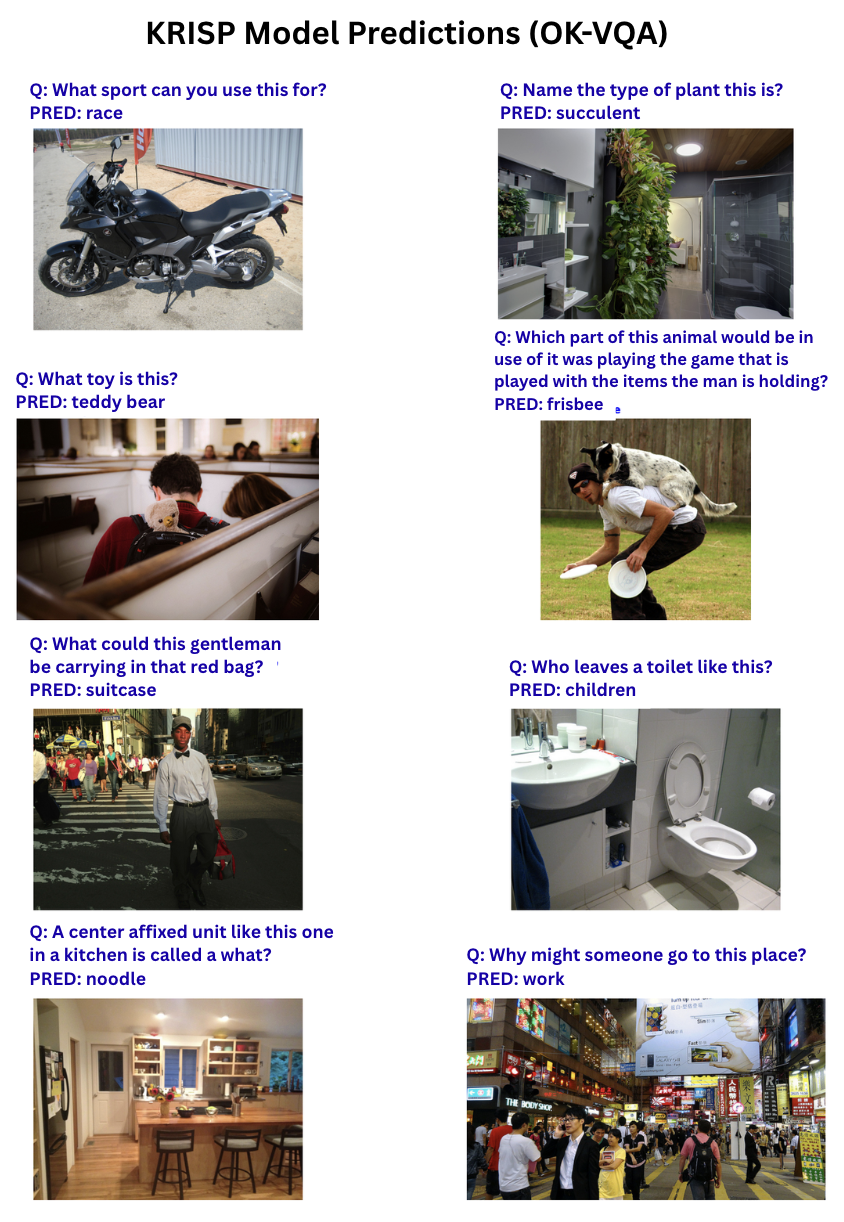}
    \caption{Reproduced KRISP Sample Output}
\end{figure}

\begin{table*}[h]
\centering
\caption{Performance comparison of lightweight KRISP reproductions.}
\label{tab:results}

\resizebox{\textwidth}{!}{
\begin{tabular}{lccc}
    \toprule
    \textbf{Model} & \textbf{Trainable Params} & \textbf{Dataset} & \textbf{Accuracy} \\
    \midrule
    BASE KRISP (Ref) & 116.14M & OKVQA\cite{marino2019okvqa} & 32.37\% (RAW)\\
    \textbf{Model A} & \textbf{25.21M} & \textbf{VQAV2} & \textbf{74.14\%*} \\
    Model B   & 3.63M & DAQUAR & 27.75\%* \\
    \bottomrule
\end{tabular}
}
\vspace{3pt}
{\small *Validation accuracies are relative to the original KRISP Model.}
\end{table*}

\subsection{Training Dynamics}

\textbf{VQAV2 (Model A):} An extensive dataset utilized for critical performance assessment. The model's test accuracy was 75 \% (relative to the Base Krisp Model). Although the model converged in fewer epochs, validation loss curves indicated that the model was overfitting. Despite severe resource limitations, the validation accuracy was comparable to some prior SOTA accuracy. 

\textbf{DAQUAR Data (Model B):} From 3.12\% (epoch 1) to 9.71\% (epoch 10), training showed consistent but gradual improvement. In the last epoch, validation accuracy reached a peak of 8.88\%. The difficult nature of DAQUAR's extensive answer vocabulary (582 classes) and intricate spatial reasoning questions is reflected in the low absolute accuracy. Notable trends:

\begin{enumerate}
    \item The model exhibits bias in favor of common responses (e.g., predicting "table", "chair", "2").
    \item Compared to counting or color queries, object existence questions perform better.
    \item Occasional correct spatial reasoning (e.g., "cabinet" for "object on floor right of cabinet")
\end{enumerate}

\subsection{Knowledge Retrieval Analysis}
Analysis of retrieved ConceptNet triples revealed:
\begin{enumerate}
    \item Relevant objects are successfully identified by image-grounded retrieval (e.g., for kitchen images: "room /r/Antonym kitchen", "monitor /r/AtLocation desk").
    \item Antonym relationships dominate retrieved triples, which may provide limited semantic value
    \item For each question-image pair, an average of three to four pertinent triples were retrieved.
    \item The accuracy of CLIP's object detection\cite{minderer2022simple} is correlated with retrieval quality.
\end{enumerate}

\section{Discussion}

\subsection{Performance Gap Analysis}
Although Model A's performance (24\%) in the VQAV2 \cite{goyal2017making} dataset was below the base model KRISP reference, it still has a good chance of hitting the SOTA threshold.
The architecture \textit{can} learn when the task complexity is suitably scaled, as shown by the proof-of-concept on synthetic data (66.57\%).

\subsection{Impact of Knowledge Integration}
The image-grounded knowledge retrieval shows promise but faces limitations:
\begin{enumerate}
   \item \textbf{Quality vs. Quantity:} ConceptNet's extensive coverage is accompanied by noise; a large number of retrieved triples are not semantic enrichments but rather tangentially related antonyms.
    \item \textbf{Grounding Assists:} Image-conditioned retrieval avoids entirely irrelevant information, but it is unable to address basic problems with data quality.
    \item \textbf{}{} {Attention Weights:} The model demonstrates that the fusion mechanism functions as intended by learning to selectively pay attention to knowledge.
\end{enumerate}

\subsection{Scalability Challenges}
Our reproduction efforts confirm that KRISP's conceptual design is sound, but its effectiveness heavily depends on:
The conceptual design of KRISP is sound, according to our reproduction efforts, but its efficacy largely depends on:
\begin{enumerate}
\item \textbf{Scale:} Strong performance seems to require large backbone models and extensive pretraining. 
\item \textbf{Knowledge Quality:} Filtering or task-specific curation may be necessary for Raw ConceptNet.
\item \textbf{Answer Space:} As vocabulary grows, performance deteriorates quickly (26 classes → 582 classes).

\end{enumerate}

Although there is a minimum scale threshold below which reasoning capabilities collapse, the fact that our lightweight model achieves approximately 75 \% of the original KRISP performance with 21.7 \% of parameters suggests the core mechanism is parameter-efficient.

\subsection{Architectural Insights}

When compared to direct concatenation, the two-stage attention mechanism is advantageous because: 
\begin{enumerate}
  \item Maintaining visual grounding through separate image-question fusion
    \item Allowing selective knowledge integration via second attention stage
    \item Preventing knowledge features from dominating the representation space
\end{enumerate}

Nevertheless, despite these advancements, the model still has trouble answering intricate spatial reasoning and fine-grained attribute questions, indicating that architectural innovation by itself is unable to make up for inadequate model capacity.

\section{Conclusion}
We reexamine KRISP from a lightweight standpoint and highlight the scalability issues and implicit engineering assumptions underlying knowledge-enhanced VQA systems. Using only about 20\% of the original KRISP model’s computational size, we achieve approximately 75\% of the SOTA accuracy. Furthermore, by experimenting with an evolved architecture on real-world DAQUAR evaluation, we demonstrate through systematic ablation studies that:

\begin{enumerate}
   \item The fundamental idea of knowledge-grounded reasoning in KRISP is both parameter-efficient and architecturally sound, though it could be more computationally efficient.
  \item  While image-grounded knowledge retrieval increases relevance, it is unable to overcome noisy knowledge bases.
  \item For complex VQA, extreme parameter reduction  results in fundamental capacity bottlenecks. 
  \item  When integrating external knowledge, two-stage attention fusion aids in preserving visual grounding.

\end{enumerate}

Our partial reproduction offers a benchmark for future simplifications of knowledge-augmented VL models and sheds light on how structure-grounded reasoning responds to extreme model compression. To close the gap between lightweight and full-scale implementations, future research should investigate knowledge base curation, intermediate-scale models, and hybrid retrieval techniques.

\section{Limitations}
\begin{enumerate}
   \item Due to computational limitations, there is no direct comparison with the original KRISP on the same datasets. 
  \item ConceptNet quality constraints (antonym-heavy triples) are not addressed.Longer training may yield better results.
  \item DAQUAR evaluation is limited to 10 epochs.
 \item  Synthetic VQA item Proof-of-concept might not accurately represent the complexity of real-world tasks.{itemize}
\end{enumerate}

\bibliographystyle{plain}   
\bibliography{references}

\end{document}